\documentclass[10pt,twocolumn,letterpaper]{article}

\usepackage{cvpr}
\usepackage{times}
\usepackage{epsfig}
\usepackage{graphicx}
\usepackage{amsmath}
\usepackage{amssymb}
\usepackage{microtype}
\usepackage{multirow}
\usepackage{graphicx}
\usepackage{subfigure}
\usepackage{threeparttable}
\usepackage{booktabs}
\usepackage{array}

\usepackage[pagebackref=true,breaklinks=true,letterpaper=true,colorlinks,bookmarks=false]{hyperref}

\cvprfinalcopy 

\begin{document}

\title{Seeing Convolution Through the Eyes of Finite Transformation Semigroup Theory: An Abstract Algebraic Interpretation of Convolutional Neural Networks}

\author{Andrew Hryniowski$^{1,2,3}$, Alexander Wong$^{1,2,3}$\\
			$^{1}$ Video and Image Processing Research Group, Systems Design Engineering, University of Waterloo\\
			$^{2}$ Waterloo Artificial Intelligence Institute, Waterloo, ON\\
			$^{3}$ DarwinAI Corp., Waterloo, ON\\ 			
			\texttt{$\{$apphryni, a28wong$\}$@uwaterloo.ca}
	}

\maketitle

\begin{abstract}
Researchers are actively trying to gain better insights into the representational properties of convolutional neural networks for guiding better network designs and for interpreting a network's computational nature. Gaining such insights can be an arduous task due to the number of parameters in a network and the complexity of a network's architecture. Current approaches of neural network interpretation include Bayesian probabilistic interpretations and information theoretic interpretations.  In this study, we take a different approach to studying convolutional neural networks by proposing an abstract algebraic interpretation using finite transformation semigroup theory. Specifically, convolutional layers are broken up and mapped to a finite space. The state space of the proposed finite transformation semigroup is then defined as a single element within the convolutional layer, with the acting elements defined by surrounding state elements combined with convolution kernel elements. Generators of the finite transformation semigroup are defined to complete the interpretation.  We leverage this approach to analyze the basic properties of the resulting finite transformation semigroup to gain insights on the representational properties of convolutional neural networks, including insights into quantized network representation.  Such a finite transformation semigroup interpretation can also enable better understanding outside of the confines of fixed lattice data structures, thus useful for handling data that lie on irregular lattices. Furthermore, the proposed abstract algebraic interpretation is shown to be viable for interpreting convolutional operations within a variety of convolutional neural network architectures. 

\end{abstract}
\begin{figure*}
    \centering
    \includegraphics[width=1.0\linewidth]{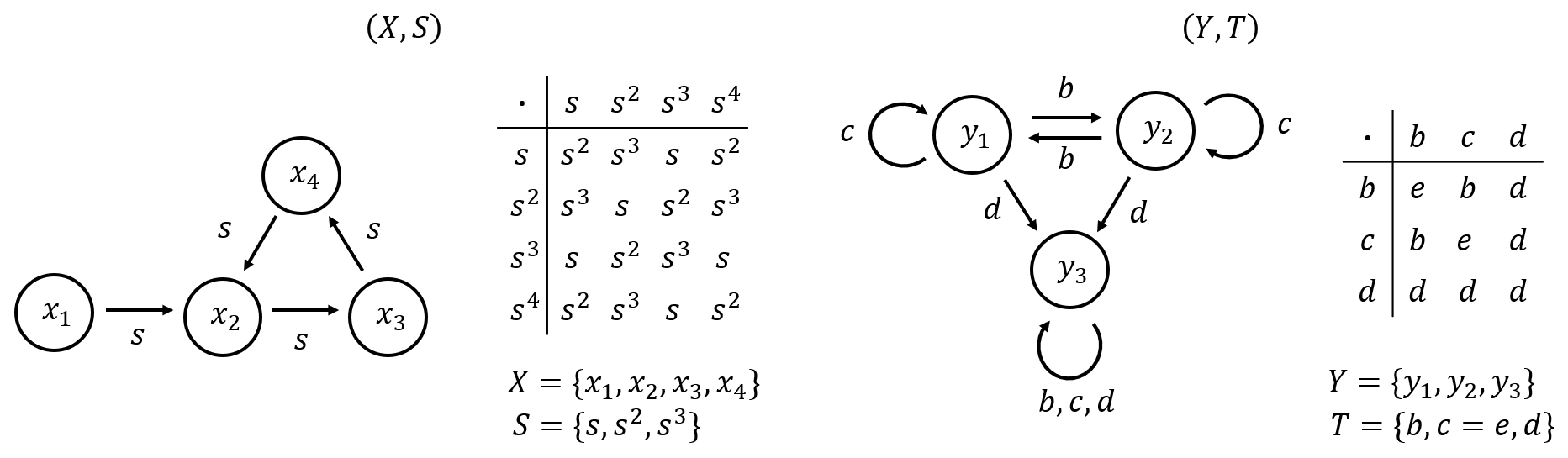}
    \caption{\small Two examples of finite transformation semigroups. Each transformation semigroup is depicted as an automaton in conjunction with the corresponding semigroup multiplication table. The finite transformation semigroup $(X,S)$ on the left is generated by one element $a$. The finite transformation semigroup $(Y,T)$ on the right is generated by three elements $b$, $c$, and $d$. Both $S$ and $T$ consist of three unique functions.  }
    \label{fig:semi_example}
\end{figure*}

\section{Introduction}
Convolutional neural networks (CNNs)~\cite{lecun1998gradient} have demonstrated tremendous success in recent years for a large range of applications, particularly for prediction using structured data. Despite such successes, a major challenge with leveraging convolutional neural networks is the sheer number of learnable parameters within such networks, making understanding and gaining insights about them a daunting task. As such, researchers are actively trying to gain better insights and understanding into the representational properties of convolutional neural networks, especially since it can lead to better design and interpretability of such networks.  

One direction that holds a lot of promise in improving understanding of convolutional neural networks, but is much less explored than other approaches, is the construction of theoretical models and interpretations of such networks.  Current approaches of neural network interpretation include Bayesian probabilistic interpretations~\cite{Polson} and information theoretic interpretations~\cite{Zhao,Tishby15,Tishby}.  Such theoretical models and interpretations could help guide and motivate design decisions of convolutional neural networks.  

Motivated by this direction, in this study we take a different approach to interpreting and gaining insight into the nature of convolutional neural networks through the use of finite transformation semigroup theory~\cite{semigroup}. To achieve this goal, we introduce an abstract algebraic interpretation of convolutional operations enabled by a novel approach for the interpretation of convolutional layers in convolutional neural networks as finite transformation semigroups. This allows us to study the basic properties of the resulting finite transformation semigroup from an abstract algebraic interpretation to gain insights on the representational properties of convolutional layers. 

The remainder of this paper is organized as follows. First, Section~\ref{sec:background} reviews convolution, discusses challenges of network analysis, introduces efforts into neural network quantized representation, and defines the notion of finite transformation semigroups. Then, Section~\ref{sec:dnn_semi} proposes a novel technique for interpreting convolutional layers within convolutional neural networks as a finite transformation semigroup. Next, Section~\ref{sec:properties} analyzes properties of the proposed abstract algebraic interpretation of convolutional neural networks. Section~\ref{sec:experiment} studies the effect of using the proposed abstract algebraic interpretation with different number of states in the finite transformation semigroups to model convolution operations and the associated effects on network performance. Finally, Section~\ref{sec:dicussion} summarizes the results and suggests future areas of research. 

\section{Background}
\label{sec:background}

In this section, we will review convolutions in the context of convolutional neural networks, discuss the challenges of analyzing convolutional neural networks, discuss efforts into neural network quantized representation, and define mathematically the notion of finite transformation semigroups.

\subsection{Convolutions in Convolutional Neural Networks}

Convolutions are not only a key aspect of convolutional neural networks, but also the major computational work horse of such networks.  As such, a better understanding of what they are computing, and how they can be represented and analyzed would allow for a host of interesting insights that can drive better design, ranging from more efficient network architecture designs and representations~\cite{lin2016fixed,courbariaux2016binarized,Jacob,howard2017mobilenets, shafiee2018deep, wong2018ferminets,ShuffleNetv1,AttoNets} to new network architectures with greater representational capacity for improved accuracy~\cite{he2016deep, zoph2016neural}. 

Convolutional neural networks are primarily built from a series of stacked convolutional layers, each of which are parameterized by sets of weights $w^i$ contained on some lattice structure. The lattice structure of the weight is dependent on the type of data the convolutional neural network is designed for. Convolution operations can be described by
\begin{equation}
    \label{eq:conv}
    x^{i+1}_k = \sum_{ j \in |c_i| } x^i_j * w^i_{jk}
\end{equation}
where $i$ is the index of the current layer in a given network, $j$ is the index of the input channel being operation on, $k$ is the index of the output channel, $x^i_j$ is the feature map being operated on, $|c_i|$ is the number of input feature maps, and $x^{i+1}_k$ is the output feature map. Note that a bias term, which is typically added to then entire output channel, is omitted from Equation~\ref{eq:conv} and is not considered as part of convolutional operations in this work.  Other convolution  techniques include dilated convolution~\cite{yu2015multi}, depthwise separable convolution~\cite{howard2017mobilenets}, and rotation equivariant convolution~\cite{veeling2018rotation}. 

\subsection{Network Analysis}

Convolutional neural networks in general are quite difficult to analyze. With an enormous number of trainable parameters it can be extremely difficult to determine what components of an input led to any given prediction, and more difficult still to determine what features a network has learned to detect.  Some convolutional neural networks are built for data that can not be easily mapped to a regular lattice. The lack of well-defined regular lattice structures in the data representation increases the difficulty of analyzing its internal behavior. An issue with such convolutional neural networks is that the network structure can only be applied to a specific graph structure that is designed for a particular instantiation of the data being analyzed~\cite{scarselli2009graph, defferrard2016convolutional}. Generalized analysis techniques that can be applied to analyze heterogeneous network structures are highly desired as they would allow for greater insights into network behavior. 

Some current approaches of neural network interpretation include Bayesian probabilistic interpretations~\cite{Polson} and information theoretic interpretations~\cite{Zhao,Tishby15,Tishby}. For example, a Bayesian interpretation was explored in~\cite{Polson} from the perspective of Kolmogorov’s representation of a multivariate response surface, where operations within a neural network can be seen as a superposition of univariate activation functions applied to an affine transformation of the input variable.  In~\cite{Tishby15,Tishby}, an information theoretical interpretation of neural networks was explored where networks are quantified by the mutual information between layers in the network as well as the input and output variables. Further investigation into alternative interpretations can lead to new insights beyond what these existing interpretations can provide.

\subsection{Neural Network Quantized Representation}

The precision of data representation of the weights of a convolutional neural network can have a significant impact on the size and computational complexity of inference of a convolutional neural network.  While training is typically conducted on convolutional neural networks with floating-point data representations of weights, it has been shown that such convolutional neural networks can still achieve strong inference performance using quantized representations of  weights~\cite{lin2016fixed}, even when data precision is reduced all the way down to 1 bit~\cite{courbariaux2016binarized}. It has also been shown that the more one quantizes the weights of a convolutional neural network, the greater the impact on modeling performance~\cite{lin2016fixed}. Quantization of a network can occur after a network has been trained~\cite{gong2014compressing}, allowing the quantization to take use-case and hardware considerations into account. Another approach builds network quantization directly into the training process in order to minimize the performance loss incurred from the quantization procedure~\cite{courbariaux2016binarized, gupta2015deep}. As such, having a better analytic understanding into the effect and tradeoffs of various levels of quantized representation can lead to better decisions on network representation design.

\subsection{Finite Transformation Semigroups}

Convolutional operations in a convolutional neural network can be viewed as mapping information from one state to another. A convolutional neural network is required to learn a variety of operations to perform effectively. Studying the nature of these operations is integral to understand what a network is computing. Formalizing these operations within an algebraic framework would allow theoretical analysis to more easily be applied. Abstract algebra can serve as useful tool in this endeavour.

Part of abstract algebra is the study of associative systems called Finite Transformation Semigroups~\cite{semigroup}. A finite transformation semigroup $(X, S)$ is defined by a finite set of states $X=\{x_1, \dots, x_m\}$ and a finite semigroup $S = \{s_1, \cdots, s_t\}$. The finite semigroup $S$ is a set of maps (functions) between states in $X$. All state mappings must be closed under composition $X \cdot S \subseteq X$. The semigroup $S$ must satisfy two properties, closure $SS \subseteq S$, and associativity 
\begin{equation}
    x_i\cdot (s_j s_k) = (x_i \cdot s_j)\cdot s_k
\end{equation} 
where $x_i \in X$ and $s_j,s_k \in S$. These properties allow a semigroup to be represented using a multiplication table. A semigroup can be generated by a subset of its functions $S=\langle \{s_p\} \rangle$ where $\{s_p\} \in S$, and $1 \leq p \leq t$. All elements in $S$ are words formed by the generating elements. Note that the size of a semigroup is the number of unique functions, and not the number of generating elements. Two examples of finite transformation semigroups are shown in Figure~\ref{fig:semi_example}. 

Two special types of semigroups are monoids and groups. A monoid  is a semigroup with an identity map $e$. An identity maps all states to themselves. All identity maps in a monoid  are equivalent. A group is a monoid  where all maps are reversible. That is, for every element $g_i \in G$, a group, there exists a corresponding element $g_j \in G$ such that 
\begin{equation}
   g_i g_j = g_i g_i^{-1} = e
\end{equation} 
where $g_i^{-1}=g_j$ is the inverse of $g_i$. Groups are often used to study the symmetries of systems~\cite{semigroup}.

One type of analysis of a finite transformation semigroup is decomposing it into \textit{irreducible components} (atomic elements) of simple groups and flip-flops (i.e., two-element right-zero monoids)~\cite{rhodes2010applications}. Decomposing a semigroup in this fashion allows a coordinate system to be built using these sub-components. One such decomposition is called the holonomy decomposition~\cite{egri2015computational}. 

\section{Abstract Alegbraic Interpretation of Convolutional Neural Networks}
\label{sec:dnn_semi}

A better understanding of the functions that a convolutional layer learns would aid interpretations of convolutional neural networks and aid with network design. Here, we will introduce an abstract algebraic interpretation of convolutional neural networks via finite transformation semigroup theory. To achieve this goal, we introduce a method for interpreting convolutional layers within convolutional neural networks as finite transformation semigroups to facilitate the proposed abstract algebraic interpretation.  The details are described below. 

\subsection{Finite Space Mapping}

\begin{figure*}
    \centering
    \includegraphics[width=0.9\linewidth]{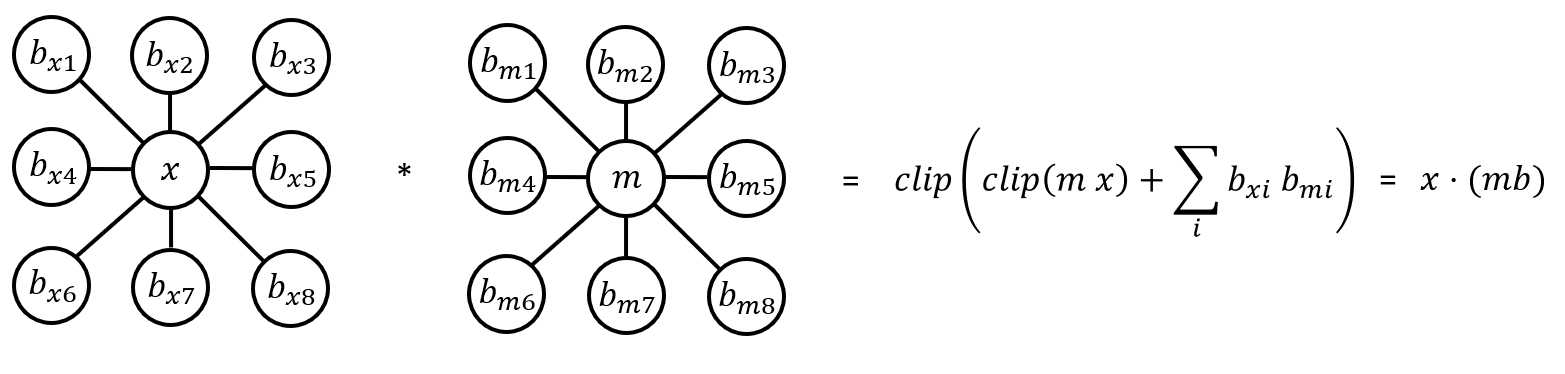}
    \vspace{0.1in}
    \caption{\small Example of mapping a  convolution operation to a semigroup action. Elements of a feature map are multiplied by elements in a convolution kernel. $x \in X$ is the state being acted upon. The semigroup elements are represented by $m$ and by the dot product of the $b$ elements from both the feature map and the convolution kernel, where $m,b,mb \in CS_r$. }
    \label{fig:conv_to_fts}
\end{figure*}

Most of the operations performed within a standard neural network are based in floating point arithmetic. To interpret convolution layers as finite transformation semigroups we must first map the floating-point state space that the convolutions acts on to a finite state space. The states that the convolutions act on are defined by the values of the input feature maps. As such, one must first establish a finite space mapping scheme for mapping neural network parameters from floating-point state space to a finite state space.

In this work, the finite space mapping scheme leveraged to map a neural network parameter to a finite space can be described as follows. Let $q_{min}$ and $q_{max}$ control the minimum values and maximum values, respectively, allowed in feature map space. All values above or below the range will be clipped to $q_{min}$ and $q_{max}$, respectively. Let $r_{min}$ and $r_{max}$ be integer values in the finite space that $q_{min}$ and $q_{max}$ are linearly mapped to, respectively. Values mapped from feature map space to finite space are rounded to the nearest integer. The same linear map is used for all convolutional layers in a convolutional neural network. Due to the nature of activation functions, the mean feature map value is often centered on or near the origin. For this reason, $r_{min}$ and $r_{max}$, and $q_{min}$ and $q_{max}$ are set to be symmetric about the origin (i.e., $-r_{min} = r_{max} = r$, $-q_{min} = q_{max} = q$).

\subsection{Finite Feature Map Elements as a State Space}

\begin{table*}
    \caption{$CS_r$ Generators}
    \vspace{0.1in}
    \centering
    \begin{tabular} { l c c c c c c c c c c c}
        \toprule
        \textbf{Current State ($x$)} & -r & -r+1 & -r+2 & $\cdots$ & -1 & 0 & 1 & $\cdots$& r-2 & r-1 & r \\  
        \midrule
   
        \textbf{Increment ($c$)} & -r+1 & -r+2 & -r+3 & $\cdots$ & 0 & 1 & 2 & $\cdots$ & r-1 & r & r \\ 

        \textbf{Zero ($z$)} & 0 & 0 & 0 & $\cdots$ & 0 & 0 & 0 & $\cdots$ & 0 & 0 & 0 \\
        
        \textbf{Identity ($e$)} & -r & -r+1 & -r+2 & $\cdots$ & -1 & 0 & 1 & $\cdots$& r-2 & r-1 & r  \\
       
        \textbf{Negative ($n$)} & r & r-1 & r-2 & $\cdots$ & 1 & 0 & -1 & $\cdots$& -r+2 & -r+1 & -r  \\
        
        \textbf{Multiplication ($m_p$)} & $p*$-r & $p*$(-r+1) & $p*$(-r+2) & $\cdots$ & $p*$(-1) & 0 & $p*$1 & $\cdots$ & $p*$(r-2) & $p*$(r-1) & $p*$r \\ [0.5ex]
        \bottomrule
    \end{tabular}
    \label{tab:basic_generators}
\end{table*}

To interpret convolutional operations as a finite transformation semigroup we need to identify the generators of the semigroup and the states they act on. One possible option for a state space is to use the entire set of feature maps. This option has two issues: i) the high dimensionality of a single feature map, and ii) the change in feature map dimensionality throughout a given network. The dimensionality of a single feature map (the input to a convolutional layer) has 3 dimensions (height, width, number of channels) and can consist of thousands of values. As an example, let us consider a feature map in a residual convolutional network ~\cite{he2016deep} with 20 layers (i.e., ResNet-20)  trained on the CIFAR-10 dataset~\cite{cifar10}. The largest feature map in this network has $32*32*16=16384$ elements, and the ResNet-20 CIFAR-10 network is already considered to a \textbf{small} network by current standards. As such, using the entire set of feature maps as the state space is computationally impractical. In addition, changing the feature map size would require allowing \textit{reshaping} elements in the semigroup, or to zero pad smaller feature maps to match larger feature map sizes, thus greatly increasing the complexity of the interpretation. 

Another option for the state space is to use the individual components of the input feature maps. The individual components are single value elements of the feature map matrix. By selecting the state space as such we are modeling the finest scale of functions in the network that are used to construct more complex functions. Modelling these \textit{building block} functions also allows the changing characteristics throughout a network to be analyzed. For example, in a convolutional neural network designed for visual perception, one can leverage the aforementioned approach to study and analyzed whether the functions that detect lines and edges in the lower levels of a network are similar to the functions that detect object level abstractions near the top level of a network. 

Using the first option allows for a better view of how your convolutional neural network is behaving from a input sample perspective. That is, determine what computations are being used to detect feature specific operations. In addition, one could investigate the inter-dependencies between the nature of computation and specific feature patterns. However, this option is simply too computationally expensive. On the other hand, the second option of using the feature map components as the state space allows for fine grained analysis of the functions that a network learns to detect features. In addition, this option is computationally inexpensive, relatively speaking, as each state is a one element vector. For this study, feature map components will be used as the state space of the finite transformation semigroup.

\subsection{Convolutions as Semigroup Actions}
\label{sec:state_space_design}

A single value in an output feature map requires $2*kh_i*kw_i$ values from the input feature map, where $kh_i$ and $kw_i$ are the height and width of kernels in layer $i$, respectively. $kh_i*kw_i$ elements come from a feature map, and the other $kh_i*kw_i$ elements come from a convolutional kernel. Note that $2*kh_i*kw_i$ assumes a regular 2-dimensional lattice, other lattices may be used but are not explored in this work. Mapping $2*kh_i*kw_i$ values to a single value is in conflict with finite transformation semigroup theory as the functions within the semigroup must map a single state to another state. The convolution operation as a whole maps a combination of states to a single state in a different space.  In the proposed abstract algebraic interpretation of convolutional neural networks, the convolution operation will be broken up into sub-convolution operations. These sub-operations are when a single lattice of learned parameter is applied to an subset of features within a feature map channel. The center value in the lattice is taken to be the state for which the semigroup action is being applied. Let this state be denoted as $x^i_{js}$, where $i$ is the layer within a network, $j$ is the channel index in layer $i$, and $s$ is the element index within channel $j$. The remaining $2*kh_i*kw_i-1$ elements will define the properties of the action. The convolution operation can be interpreted as a finite linear operation of the form
\begin{equation}
    \label{eq:linear}
    f(\hat{x}^i_{js}, m, b) = clip(clip(m \hat{x}^i_{js})+b)
\end{equation}
where $\hat{x}^i_{js}$ is the quantized state being acted on, $m$ and $b$ are representations of the semigroup action, and $clip$ bounds the result to plus-minus $r$. In this context, $\hat{x}^i_{js}$ is the state, $m$ is a multiplicative action on $\hat{x}^i_{js}$, and $b$ is an additive (or subtractive) action on the result of the first action. In multiplicative semigroup notation, $f(\cdot)$ can be written as
\begin{equation}
    \label{eq:lin_action}
    f(\hat{x}^i_{js}, m, b) = \hat{x}^i_{js} \cdot m \cdot b = \hat{x}^i_{js} \cdot ( m  b )
\end{equation}
where $m$ and $b$ are actions representing possible quantized multiplication operation and quantized addition operations, respectively. Figure~\ref{fig:conv_to_fts} demonstrates an example of mapping a convolution operation to a semigroup action.

\subsection{Semigroup Generators}

Let $(X_r, CS_r)$ represent the proposed finite transformation semigroup where $X_r = \{-r,\dots,0,\dots,r\}$ is the state space with $2r+1$ states, and $CS_r$ is the semigroup acting on $X_r$. The generators of $CS_r$ used to model convolution operations are a set of four base generators required for all $r \geq 1$, and $|M_r|$ multiplication generators where $M_r$ is the set of prime multiplications generators in $(X_r, CS_r)$. 

\begin{figure*}
    \centering
    \includegraphics[width=1.0\linewidth]{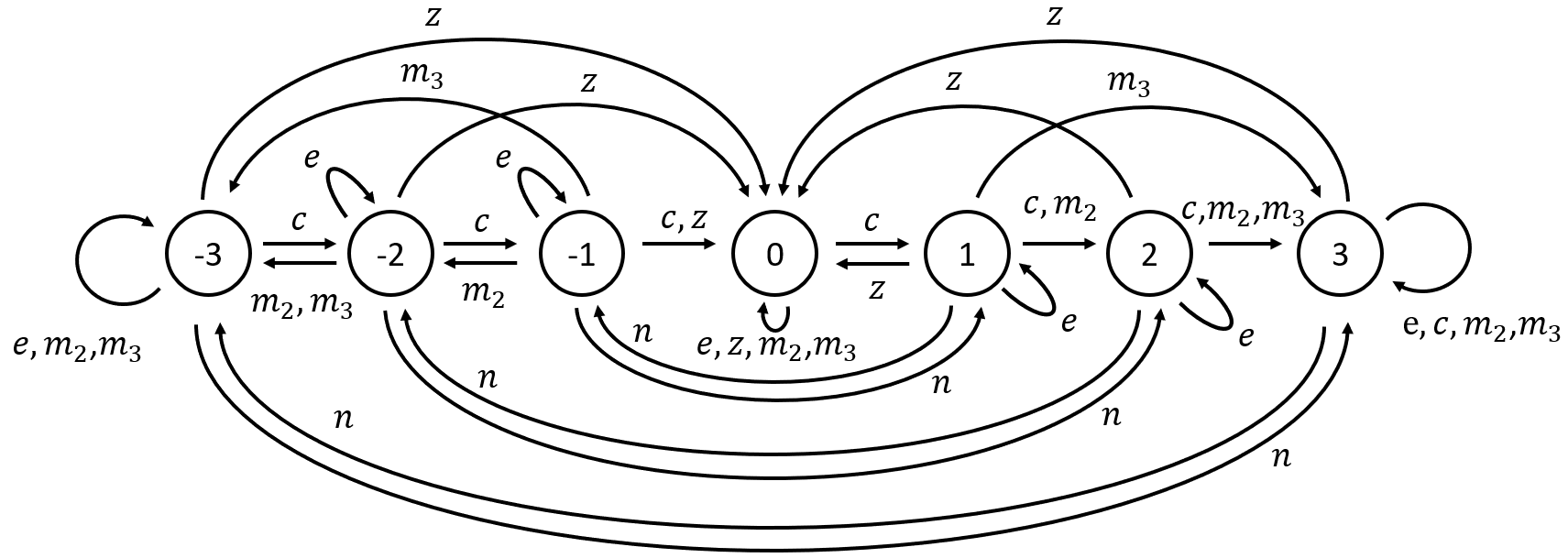}
    \vspace{0.1in}
    \caption{\small An example of the proposed finite transformation semigroup $CS_3$ in automata form. $CS_3$ is generated by 6 generators; the 4 basic generator and the 2 prime generators, $m_2$ and $m_3$. }
    \label{fig:conv_example}
\end{figure*}

The types of generators in the semigroup are shown in Table~\ref{tab:basic_generators}. The increment generator $c$ is adding one to all states, the zero generator $z$ maps all states to the origin, the identity generator $e$ is a \textit{do nothing} operation, and the negative generator $n$ symmetrizes states about the origin. Note that the identity element $e$ makes $CS_r$ a monoid. In addition, a decrement operation can be formed by $n c n$. $m_p$ is the multiplication generator for some number $p$. The only required $p$'s are prime numbers where $p\in \mathbb{N}$, $ 1 < p \leq r$. An example automata of $r=3$ is shown in Figure~\ref{fig:conv_example}. Table~\ref{tab:cs_1_mul_table} shows an example of $CS_1$ in multiplication table format.

\begin{table}
    \caption{Number of Generators and Size of $CS_r$ }
    \vspace{0.1in}
    \centering
    \begin{tabular} { l c c c c }
        \toprule
        \textbf{r} & 1 & 3 & 7 & 15 \\  
        \midrule
        \textbf{Bits} & 2 & 3 & 4 & 5  \\ 
        \textbf{Generators} & 4 & 6 & 8 & 10   \\ 
        \textbf{Size} & 13 & 719 & 30139 & 1122143 \\
        \bottomrule
    \end{tabular}
    \label{tab:num_gen}
\end{table}

\begin{figure}
    \centering
    \includegraphics[width=1\linewidth]{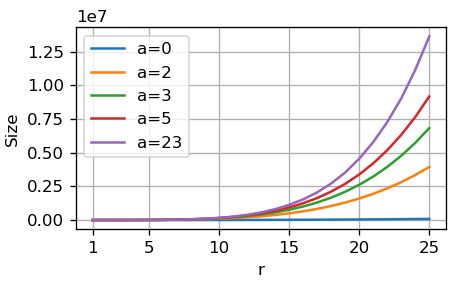}
    \caption{\small The number of elements in $CS_{r-a}$ for $a={0,2,3,5,23}$. At $a=23$ all prime generators are used for all $r \leq 25$. }
    \label{fig:conv_prime_reduction_partial}
\end{figure}

\begin{table*}
    \centering
    \caption{The multiplication table for $CS_1$. $CS_1$'s generating set is $\langle c, e, n, z \rangle$ and contains 13 unique functions.}
    \vspace{0.1in}
    \begin{tabular}{c|ccccccccccccc}
   $\cdot$ & $ccn$ & $ncn$ & $cncn$ & $e$ & $cn$ & $ncncn$ & $z$ & $ncnc$ & $c$ & $n$ & $cnc$ & $nc$ & $cc$ \\ 
        \hline             
         $ccn$ & $ccn$ & $ccn$ & $ccn$ & $ccn$ & $z$ & $z$ & $z$ & $z$ & $z$ & $cc$ & $cc$ & $cc$ & $cc$ \\
         $ncn$ & $ccn$ & $ccn$ & $ncn$ & $ncn$ & $ncncn$ & $z$ & $z$ & $z$ & $ncnc$ & $nc$ & $nc$ & $cc$ & $cc$ \\
         $cncn$ & $ccn$ & $ccn$ & $cncn$ & $cncn$ & $cn$ & $z$ & $z$ & $z$ & $c$ & $cnc$ & $cnc$ & $cc$ & $cc$ \\ 
         $e$ & $ccn$ & $ncn$ & $cncn$ & $e$ & $cn$ & $ncncn$ & $z$ & $ncnc$ & $c$ & $n$ & $cnc$ & $nc$ & $cc$ \\ 
         $cn$ & $ccn$ & $ccn$ & $cn$ & $cn$ & $cncn$ & $z$ & $z$ & $z$ & $cnc$ & $c$ & $c$ & $cc$ & $cc$ \\
         $ncncn$ & $ccn$ & $ccn$ & $ncncn$ & $ncncn$ & $ncn$ & $z$ & $z$ & $z$ & $nc$ & $ncnc$ & $ncnc$ & $cc$ & $cc$ \\  
         $z$ & $ccn$ & $ccn$ & $z$ & $z$ & $ccn$ & $z$ & $z$ & $z$ & $cc$ & $z$ & $z$ & $cc$ & $cc$ \\
         $ncnc$ & $ccn$ & $ncn$ & $z$ & $ncnc$ & $ccn$ & $ncncn$ & $z$ & $ncnc$ & $cc$ & $ncncn$ & $z$ & $nc$ & $cc$ \\ 
         $c$ & $ccn$ & $cncn$ & $z$ & $c$ & $ccn$ & $cn$ & $z$ & $c$ & $cc$ & $cn$ & $z$ & $cnc$ & $cc$ \\  
        $n$ & $ccn$ & $cn$ & $ncncn$ & $n$ & $ncn$ & $cncn$ & $z$ & $cnc$ & $nc$ & $e$ & $ncnc$ & $c$ & $cc$ \\  
        $cnc$ & $ccn$ & $cn$ & $z$ & $cnc$ & $ccn$ & $cncn$ & $z$ & $cnc$ & $cc$ & $cncn$ & $z$ & $c$ & $cc$ \\ 
        $nc$ & $ccn$ & $ncncn$ & $z$ & $nc$ & $ccn$ & $ncn$ & $z$ & $nc$ & $cc$ & $ncn$ & $z$ & $ncnc$ & $cc$ \\  
        $cc$ & $ccn$ & $z$ & $z$ & $cc$ & $ccn$ & $ccn$ & $z$ & $cc$ & $cc$ & $ccn$ & $z$ & $z$ & $cc$ \\ 

    \end{tabular}
    \label{tab:cs_1_mul_table}
\end{table*}

\section{$(X_r, CS_r)$ Properties}
\label{sec:properties}

The proposed family of finite transformation semigroups $(X_r, CS_r)$ has a variable number of generators depending on the number of states in $X_r$. Here, we investigate the size of $CS_r$ as a function of $r$. An approximation of $CS_r$ is proposed to  significantly reduce the number of elements compared to $CS_r$. Then the holonomy decomposition~\cite{egri2015computational} of $(X_r, CS_r)$ is briefly explored.

\subsection{Semigroup Size}

The semigroup $CS_r$ includes an additional generator for every new prime number less than or equal to $r$. Table~\ref{tab:num_gen} shows three properties of $CS_r$: the number of generators, the semigroup size, and the number of bits required to represent the number of states in $X_r$. The number of bits required to represent the states in $X_r$ is $\lceil \log_2 (2r+1) \rceil$. The $r$'s shown in Table~\ref{tab:num_gen} reflect the largest $r$ that can be used for the corresponding number of bits. The number of generators required for a given $CS_r$ clearly increases with $r$, since all primes less than equal to $r$ are required as multiplication generators. The size of $CS_r$ increases by approximately two orders of magnitudes with every bit added over the range shown.

\subsection{Subsemigroups of $CS_r$}

$CS_r$ quickly grows as $r$ increases. Reducing the size of $CS_r$ would be beneficial to reduce the number of computations required during analysis of the semigroup. $CS_r$'s size increases as $r$ increases since the number of states in the system increases linearly with $r$, and more generators are used to construct the semigroup $CS_r$ for larger $r$. It is possible to approximate $CS_r$ by selecting a subset of the semigroup's elements. A specific type of subset, called a subsemigroup, may be formed by removing generators from $CS_r$'s definition. Let $CS_{r-a}$ be a subsemigroup of $CS_r$ (i.e., $CS_{r-a} \subseteq CS_r$) where all prime generators less than or equal to $a$ are used. If $a \geq r$ then $CS_{r-a} = CS_r$. 

Figure~\ref{fig:conv_prime_reduction_partial} compares $CS_r$'s size to its subsemigroups $CS_{r-a}$ size for $r \leq 25$ and $a=0,2,3,5,23$. For larger $r$ there is a significant reduction in number of elements. This result is surprising as the subsemigroups only differ by one generator for $a=0,2,3,5$. The increase in size of $CS_{r-a}$ diminishes with each additional generator added. The case were $a=0$ is interesting from a practical perspective. When $a=0$ the only types of multiplication actions allowed on the state are $-1$, $0$, or $1$. This subsemigroup would then require setting the center of each convolution kernel to either $-1$, $0$, or $1$. Notice that by limiting the value of only \textbf{one} element in all convolution kernels to three values the majority of all elements in $CS_r$ are wiped out.

\subsection{Transformation Semigroup Decomposition}

The size of $CS_r$ quickly grows as $r$ increases. Decomposing its structure into a coordinate system would allow for easier analysis of its structure. The holonomy decomposition~\cite{egri2015computational} is used to decompose both $CS_r$ and $CS_{r-a=0}$ for various $r$'s.  The decomposition of $CS_r$ produced a surprising result in that the only two types of building blocks of $CS_r$ are flip-flops and 2-cycle groups. This property of the decomposition is demonstrated in the generators of $CS_r$. All but two of $CS_r$'s generators act like or contain many copies of fuse like structures in that the generators, when individually applied multiple times, end up at a trivial cyclic state. Increment $c$ eventually sends all states to $r$, zero $z$ instantly \textit{kills} all states, and all multiplication generators $m_p$ eventually send all states to either $r$ or $-r$ depending on initial conditions. The negative generator $n$ is responsible for the cyclic group element in the decomposition. Applying it twice ends up at the initial state.

The coordinates generated from the decomposition of $CS_r$ are quite complex and quickly grows in depth as $r$ increases. For example, the decomposition of $CS_3$ produces a cascade semigroup with 6 generators, and has 9 levels with (4, 5, 5, 9, 5, 5, 5, 4, 3) points in each coordinate dimension, respectively. Moreover, the decomposition of $CS_3$ results multiple sets of tiles on a single level. Clearly, decomposing larger $CS_r$ results in huge coordinate systems.  

Decomposition of $CS_{r-a=0}$ produces a much cleaner and easier-to-compute deconstruction. For example, decomposition of $CS_{3-0}$ produces a cascade semigroup with 4 generators, 6 levels with (3, 3, 3, 3, 3, 3) points in each coordinate dimension, respectively. Regardless of $r$'s value, decomposition of $CS_{r-a=0}$ produces this simple coordinate system for any $a=0$.

\section{Experiments}
\label{sec:experiment}

To explore the proposed notion of studying convolutional neural networks using an abstract algebraic interpretation via finite transformation semigroup theory, we perform two experiments where we study convolutional neural networks using the proposed interpretation to gain insights into quantized network representation. 

\subsection{Experiment 1: Effects of $r$ and $q$ in $(X_r, CS_r)$ on Representational Performance}

In the first experiment, we constructed a  convolutional neural network $N$ and construct it into $(X_r, CS_r)$ interpretations with different number of states in the finite transformation semigroup. Thus allowing us to study the effect of using $(X_r, CS_r)$ interpretations at different precision levels for quantized representations of convolutional operations and the associated effects on representational performance.  

The convolutional neural network $N$ used in this first experiment is based on the LeNet-5~\cite{lecun1998gradient} convolutional neural network architecture, which leverages conventional convolutional layer configurations. Note that $relu6$ is used as the activation function. A test accuracy of 98.1\% was achieved on MNIST~\cite{lecun1998mnist}. The consequences of using $(X_r, CS_r)$ interpretations with different number of states (corresponding to different  precision levels for the convolutional neural network) on representational performance is studied by varying two parameters, the number of states in $X_r$ as determined by $r$, and the range of numbers kept $q$. Let $N_{r,q}$ denote the resulting convolutional neural network $N$ associated with $(X_r, CS_r)$ with convolution operations constrained by parameters $r$ and $q$.  The results are shown Table~\ref{tab:toy_results}. Note that individual convolution operations are operating in finite state space, but when added together may exceed the finite operating bounds.

\begin{table}[!h]
    \caption{$N_{r,q}$ Test Accuracy under $(X_r, CS_r)$}
    \vspace{0.1in}
    \centering
    \setlength\tabcolsep{4.0pt} 
    \begin{tabular} { c c | c c c c c c c}
        \toprule
        & \textbf{Bits} & 2 & 3 & 4 & 5  & 6  & 7  & 8 \\
        & \textbf{r} & 1 & 3 & 7 & 15 & 31 & 63 & 127 \\
        \midrule
        \multirow{4}{*}{\textbf{q}} 
        & 2 & 11.3 & 11.3 & 92.6 & 96.3 & 97.0 & 95.9 & 95.7 \\
        & 4 & 11.3 & 11.3 & 83.5 & 96.0 & 97.4 & 97.8 & 97.3 \\
        & 6 & 11.2 & 11.3 & 11.2 & 91.7 & 97.7 & 97.9 & 97.7 \\
        & 8 & 11.3 & 11.2 & 11.3 & 91.2 & 97.8 & 97.9 & 98.0 \\
        \bottomrule
    \end{tabular}    
    \label{tab:toy_results}
\end{table}

It can be observed that by using appropriate $r$ and $q$, $N_{r,q}$ is able to maintain the representational performance of $N$. For the same finite transformation semigroup interpretation $(X_r, CS_r)$, different levels of representational performance are achieved by varying $q$. The more states in $X_r$ (i.e., more bits of precision) the greater performance $N_{r,q}$ achieves. However, this trend does not hold for $q=2$, there is a $1.3\%$ decrease in performance between $r=31$ and $r=128$. Further investigation of the performance degradation will be required. 

As $r$ increases, there is a point at which the representational performance ceases to be random chance. Surprisingly, the increase in representational performance is a quick jump instead of a gradual increase, potentially indicating that specific computational mechanisms learned by the LeNet-5 convolutional neural network have minimal redundancies. In other words, the features that the convolutional neural network is learning are similar in nature and when one feature detector is effected by loss of precision other feature detectors are equally effected. 

Notice that for any given $r$ and $q$, the bin size resolution can be calculated. For a fixed $r$ and increasing $q$ the bin size resolution decreases and a corresponding decrease in performance is observed for smaller $r$. The performance decrease indicates that a sufficient bin size resolution (i.e., computational precision) around specific range of values is required to maintain representational performance, which can be leveraged to guide quantized representation design. 

For a constant $r$, the performance of the interpretations vary greatly based on $q$. The input to each convolution operation is the output of $relu6$ operation, except for the first convolution operation. At $q=6$ the input to a given convolution is \textit{complete} in that the range of numbers is not clipped when moving to the quantized domain due to using $relu6$. When the input is clipped to accommodate the range of states in $X_r$ (i.e., $q < 6$) the ability to maintain performance in the quantized domain is lost since less information is carried forward.  

\subsection{Experiment 2: Quantized Interpretation of Residual Convolutional Neural Network}

Guided by the observations made in the first experiment, we perform a second experiment where the convolutional neural network $N$ used is a residual convolutional neural network~\cite{he2016deep} 20 layers and preactivation (i.e., ResNet-20), trained on the CIFAR-10 dataset~\cite{cifar10} with an accuracy of 90\%. In particular, the proposed method is used to interpret the ResNet-20 network $N$ using finite transformation semigroup $(X_r, CS_r)$, thus illustrating that the proposed abstract algebraic interpretation can be viable for studying a variety of convolutional neural network architectures outside of conventional convolutional layer configurations.  

Given our observation in the first experiment that using appropriate $r$ and $q$ for the finite transformation semigroup interpretation $(X_r, CS_r)$ could enable $N_{r,q}$ to maintain representational performance in a particular quantized representation state, we choose the parameters to be $r=127$ and $q=8$ for $(X_r, CS_r)$ and explore the effects on representational performance.  It was observed that with the finite transformation semigroup interpretation $(X_{127}, CS_{127})$, where 8 bits are required to represent the states, the resulting $N_{127,8}$ has an accuracy of 89\% and thus retains the representational performance of $N$ for the most part at a lower precision level. As can be seen here, by leveraging a better understanding of design choices through the proposed abstract algebraic interpretation, one can make more informed decisions on representation choices.    

In summary, the results of this experiment, along with that of the first experiment, show that important insights can be obtained for guiding network design and representation by studying convolutional neural networks using the proposed abstract algebraic interpretation via finite transformation semigroup theory.

\section{Conclusions and Future Work}
\label{sec:dicussion}

In this study, we propose an abstract algebraic interpretation using finite transformation semigroup theory for studying and gaining insights into the representational properties of convolutional neural networks. To achieve this goal and construct such an interpretation, convolutional layers are broken up and mapped to a finite space. The state space of the proposed finite transformation semigroup is then defined as a single element within the convolutional layer, with the acting elements defined as a combination of elements that surround the state with elements of a convolution kernel. Generators of the finite transformation semigroup are defined to complete the interpretation.  The basic properties of the resulting finite transformation semigroup are then analyzed to gain insights on the representational properties of convolutional neural networks, including insights into quantized network representation.  Two experiments conducted in this study show that important insights can be obtained for guiding network design by studying convolutional neural networks using the proposed abstract algebraic interpretation using finite transformation semigroup theory.

A number of directions for future research are apparent for the proposed abstract algebraic interpretation. The first direction is using the proposed interpretation to gain insights into the behaviour of larger, more complex convolutional neural networks beyond the initial experiment performed in this study. With the proposed representation approach, it would be possible to analyze how the distribution of mapping functions change through a convolutional neural network. This may allow more feature specific detection mechanisms to be designed, and may allow better network quantization. The second research direction would address the short comings of the proposed abstract algebraic interpretation. For example, elements generated by the increment generator and the negative generator in the current interpretation represent both the effect of surrounding states and the effect of convolution kernels. Disentangling these actions may provide additional detailed insights beyond what the current interpretation can provide.

{\small
\bibliographystyle{ieee_fullname}
\bibliography{egbib}

\begin{thebibliography}{10}\itemsep=-1pt

\bibitem{courbariaux2016binarized}
Matthieu Courbariaux, Itay Hubara, Daniel Soudry, Ran El-Yaniv, and Yoshua
  Bengio.
\newblock Binarized neural networks: Training deep neural networks with weights
  and activations constrained to+ 1 or-1.
\newblock {\em arXiv preprint arXiv:1602.02830}, 2016.

\bibitem{defferrard2016convolutional}
Micha{\"e}l Defferrard, Xavier Bresson, and Pierre Vandergheynst.
\newblock Convolutional neural networks on graphs with fast localized spectral
  filtering.
\newblock In {\em Advances in neural information processing systems}, pages
  3844--3852, 2016.

\bibitem{egri2015computational}
Attila Egri-Nagy and Chrystopher~L Nehaniv.
\newblock Computational holonomy decomposition of transformation semigroups.
\newblock {\em arXiv preprint arXiv:1508.06345}, 2015.

\bibitem{gong2014compressing}
Yunchao Gong, Liu Liu, Ming Yang, and Lubomir Bourdev.
\newblock Compressing deep convolutional networks using vector quantization.
\newblock {\em arXiv preprint arXiv:1412.6115}, 2014.

\bibitem{gupta2015deep}
Suyog Gupta, Ankur Agrawal, Kailash Gopalakrishnan, and Pritish Narayanan.
\newblock Deep learning with limited numerical precision.
\newblock In {\em International Conference on Machine Learning}, pages
  1737--1746, 2015.

\bibitem{he2016deep}
Kaiming He, Xiangyu Zhang, Shaoqing Ren, and Jian Sun.
\newblock Deep residual learning for image recognition.
\newblock In {\em Proceedings of the IEEE conference on computer vision and
  pattern recognition}, pages 770--778, 2016.

\bibitem{howard2017mobilenets}
Andrew~G Howard, Menglong Zhu, Bo Chen, Dmitry Kalenichenko, Weijun Wang,
  Tobias Weyand, Marco Andreetto, and Hartwig Adam.
\newblock Mobilenets: Efficient convolutional neural networks for mobile vision
  applications.
\newblock {\em arXiv preprint arXiv:1704.04861}, 2017.

\bibitem{semigroup}
John~M Howie.
\newblock Fundamentals of semigroup theory.
\newblock {\em London Mathematical Society Monographs. New Series.}, 1995.

\bibitem{Jacob}
B. Jacob et~al.
\newblock Quantization and training of neural networks for efficient
  integer-arithmetic-only inference.
\newblock {\em arXiv:1712.05877}, 2017.

\bibitem{cifar10}
Alex Krizhevsky, Vinod Nair, and Geoffrey Hinton.
\newblock Cifar-10 (canadian institute for advanced research).

\bibitem{lecun1998mnist}
Yann LeCun.
\newblock The mnist database of handwritten digits.
\newblock {\em http://yann. lecun. com/exdb/mnist/}, 1998.

\bibitem{lecun1998gradient}
Yann LeCun, L{\'e}on Bottou, Yoshua Bengio, Patrick Haffner, et~al.
\newblock Gradient-based learning applied to document recognition.
\newblock {\em Proceedings of the IEEE}, 86(11):2278--2324, 1998.

\bibitem{lin2016fixed}
Darryl Lin, Sachin Talathi, and Sreekanth Annapureddy.
\newblock Fixed point quantization of deep convolutional networks.
\newblock In {\em International Conference on Machine Learning}, pages
  2849--2858, 2016.

\bibitem{Polson}
Nicholas~G. Polson and Vadim Sokolov.
\newblock Deep learning: A bayesian perspective.
\newblock {\em Bayesian Analysis}, 12:1275--1304, 2017.

\bibitem{rhodes2010applications}
John Rhodes, Chrystopher~L Nehaniv, and Morris~W Hirsch.
\newblock {\em Applications of automata theory and algebra: via the
  mathematical theory of complexity to biology, physics, psychology,
  philosophy, and games}.
\newblock World Scientific, 2010.

\bibitem{scarselli2009graph}
Franco Scarselli, Marco Gori, Ah~Chung Tsoi, Markus Hagenbuchner, and Gabriele
  Monfardini.
\newblock The graph neural network model.
\newblock {\em IEEE Transactions on Neural Networks}, 20(1):61--80, 2009.

\bibitem{shafiee2018deep}
Mohammad~Javad Shafiee, Akshaya Mishra, and Alexander Wong.
\newblock Deep learning with darwin: evolutionary synthesis of deep neural
  networks.
\newblock {\em Neural Processing Letters}, 48(1):603--613, 2018.

\bibitem{Tishby}
Ravid Shwartz-Ziv and Naftali Tishby.
\newblock Opening the black box of deep neural networks via information.
\newblock {\em arXiv preprint arXiv:1703.00810}, 2017.

\bibitem{Tishby15}
Naftali Tishby and Noga Zaslavsky.
\newblock Deep learning and the information bottleneck principle.
\newblock {\em arXiv preprint arXiv:1503.02406}, 2015.

\bibitem{veeling2018rotation}
Bastiaan~S Veeling, Jasper Linmans, Jim Winkens, Taco Cohen, and Max Welling.
\newblock Rotation equivariant cnns for digital pathology.
\newblock In {\em International Conference on Medical image computing and
  computer-assisted intervention}, pages 210--218. Springer, 2018.

\bibitem{AttoNets}
Alexander Wong, Zhong~Qiu Lin, and Brendan Chwyl.
\newblock Attonets: Compact and efficient deep neural networks for the edge via
  human-machine collaborative design.
\newblock {\em arXiv preprint arXiv:1903.07209}, 2019.

\bibitem{wong2018ferminets}
Alexander Wong, Mohammad~Javad Shafiee, Brendan Chwyl, and Francis Li.
\newblock Ferminets: Learning generative machines to generate efficient neural
  networks via generative synthesis.
\newblock {\em arXiv preprint arXiv:1809.05989}, 2018.

\bibitem{yu2015multi}
Fisher Yu and Vladlen Koltun.
\newblock Multi-scale context aggregation by dilated convolutions.
\newblock {\em arXiv preprint arXiv:1511.07122}, 2015.

\bibitem{ShuffleNetv1}
Xiangyu Zhang, Xinyu Zhou, Mengxiao Lin, and Jian Sun.
\newblock Shufflenet: An extremely efficient convolutional neural network for
  mobile devices.
\newblock In {\em arXiv:1707.01083}, 2017.

\bibitem{Zhao}
Tianchen Zhao.
\newblock Information theoretic interpretation of deep learning.
\newblock {\em arXiv preprint arXiv:1803.07980}, 2018.

\bibitem{zoph2016neural}
Barret Zoph and Quoc~V Le.
\newblock Neural architecture search with reinforcement learning.
\newblock {\em arXiv preprint arXiv:1611.01578}, 2016.

\end{thebibliography}
}

\end{document}